# Visualizing attention zones in machine reading comprehension models

Yiming Cui,[1,2,3,4,*] Wei-Nan Zhang,[1] and Ting Liu[1]

[1]Research Center for Social Computing and Information Retrieval (SCIR), Harbin Institute of Technology, Harbin 150001, China

[2]State Key Laboratory of Cognitive Intelligence, iFLYTEK Research, Beijing 100083, China

[3]Technical contact

[4]Lead contact

*Correspondence: ymcui@ir.hit.edu.cn

## Summary

The attention mechanism plays an important role in the machine reading comprehension (MRC) model. Here, we describe a pipeline for building an MRC model with a pretrained language model and visualizing the effect of each attention zone in different layers, which can indicate the explainability of the model. With the presented protocol and accompanying code, researchers can easily visualize the relevance of each attention zone in the MRC model. This approach can be generalized to other pretrained language models.

For complete details on the use and execution of this protocol, please refer to Cui et al. (2022).

## Highlights

- Visualization of different attention zones in MRC models
- Steps for training an English MRC system using SQuAD dataset and pre-trained BERT model
- Detailed quantitative analysis on the raw data to allow robust experimentation

## Graphical abstract

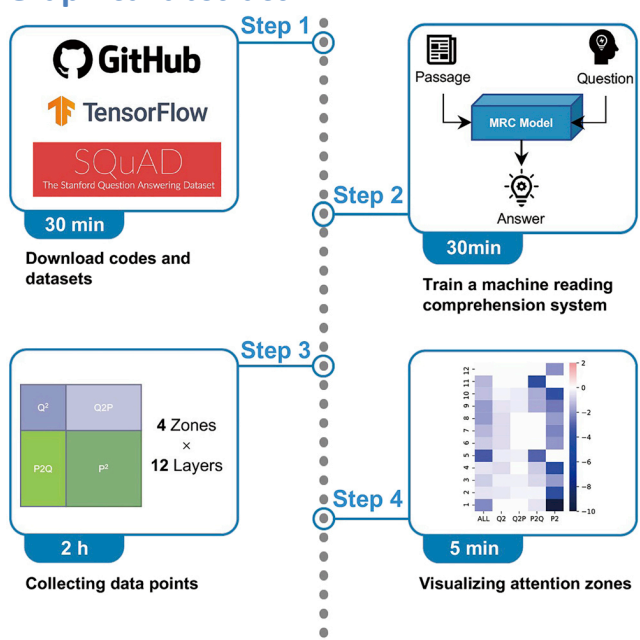

## Before you begin

One of the important tasks in artificial intelligence (AI) is to read and comprehend human language. Machine reading comprehension (MRC) is a task to achieve such goals, requiring the machine to read a passage and answer the question. With the development of deep learning techniques, we have seen rapid progress in building effective machine reading comprehension systems. After the emergence of the pre-trained language model (PLM), such as BERT (Devlin et al., 2019), which learns text semantics from large-scale text corpora, some of the MRC systems can achieve human-level performance in several MRC benchmarks.

However, a drawback of the deep learning approach is that it is difficult to explain its internal mechanism, raising concerns about building trustful and reliable AI systems. Regarding the explainability of MRC systems, we wonder how these models comprehend human language and answer the questions. An important component in the MRC system is the attention-based approach (Bahdanau et al., 2015), which explicitly assigns the "important score" for each input token. For the MRC system built on PLM, the core component is the multi-head self-attention mechanism (Vaswani et al., 2017), which increases accuracy more than vanilla attention. However, whether all attention values have significant impacts on the final system performance is uncertain.

In this context, we propose to visualize the attention by using a multilingual and multi-aspect way to comprehensively understand whether these attentions can be explainable (Cui et al., 2022). Instead of analyzing the attention matrix as a whole, we decompose the attention matrix into four different attention zones to explicitly analyze their behaviors. This protocol illustrates the main approach that was used in our previous study. It demonstrates a step-by-step method of building an MRC system, collecting data points, and visualizing the attention zones. This helps us better understand which part of attention zones is important for language comprehension.

We describe the preliminary works before we perform actual experiments and visualization. These include hardware and system requirements and package installation procedures.

### Hardware and system

Timing: 2 min

The presented protocol is mainly developed under a computer server with the Linux operation system (Debian GNU/Linux 9). The server is equipped with an Intel(R) Xeon(R) CPU @ 2.80 GHz and has 16G system memory, which is operated by Google Compute Engine. The primary computing device for the training model is a cloud TPU v2-8 with 64G high-band memory (HBM). However, the presented protocol may also be applicable to a wide range of other Linux distributions, CPUs, and GPUs.

- To start with our experiment, we need to create a TPU with the following command.

```
> ctpu up --tpu-only --zone us-central1-f --name t2-1 --tf-version 1.15.3 --tpu-size v2-8
```

This command will create a TPU (v2-8) named "t2-1" in the "us-central1-f" zone, equipped with TensorFlow 1.15.3. Please keep these in mind, as we will use this meta-data for configuring our training script.

Critical: To use a TPU, the server should be operated by Google Compute Engine. It is not possible to create TPU with other types of servers. However, there are no such restrictions if you are using GPU or CPU for training the model and can simply ignore the configurations related to TPU.

Note: We understand that some users may use macOS or Windows operating system (OS) with CPU or GPU for conducting this protocol. To help these users, we also provide several hints (non-exhaustive) on how to use our protocol under these systems. Please see troubleshooting 1 for more information.

### Package installation

Timing: 5 min

Before starting the experiment, it is necessary to install the Python dependencies. Our experiments are carried out under the Python 3.7.3 environment, but it is flexible to use other versions of Python unless it is under version 3.X and compatible with the following python dependencies. Note that it is recommended to use the "pip" command to automatically install the packages recursively. To train a machine reading comprehension model, we need to install TensorFlow (Abadi et al., 2016). The visualization requires matplotlib (Hunter, 2007) and seaborn (Waskom, 2021) libraries. Please see the solutions in troubleshooting 2 when the installation fails.

```
> pip install tensorflow==1.15.3
> pip install matplotlib
> pip install seaborn
```

## Key resources table

| REAGENT or RESOURCE | SOURCE | IDENTIFIER |
|---|---|---|
| Deposited data | | |
| SQuAD Training Set | (Rajpurkar et al., 2016) | https://rajpurkar.github.io/SQuAD-explorer/dataset/train-v1.1.json |
| SQuAD Dev Set | (Rajpurkar et al., 2016) | https://rajpurkar.github.io/SQuAD-explorer/dataset/dev-v1.1.json |
| Software and algorithms | | |
| Python | Python Software Foundation | https://www.python.org |
| TensorFlow | (Abadi et al., 2016) | https://tensorflow.org |
| matplotlib | (Hunter, 2007) | https://matplotlib.org |
| seaborn | (Waskom, 2021) | https://seaborn.pydata.org |
| MRC Model Analysis | (Cui et al., 2022) | GitHub: https://github.com/ymcui/mrc-model-analysis Zenodo: https://doi.org/10.5281/zenodo.6522993 |
| Other | | |
| English BERT-base-cased | (Devlin et al., 2019) | https://storage.googleapis.com/bert_models/2018_10_18/cased_L-12_H-768_A-12.zip |

## Step-by-step method details

### Data and code preparation

Timing: 30 min

In this protocol, we use the well-known machine reading comprehension (MRC) dataset, SQuAD (Rajpurkar et al., 2016), and English BERT (Devlin et al., 2019) (base-cased version) pre-trained language model to train an MRC system.

1. Download source codes.
   a. The source codes can be downloaded by the following command.
      ```
      > git clone https://github.com/ymcui/mrc-model-analysis
      ```
   b. Navigate into the downloaded folder.
      ```
      > cd mrc-model-analysis
      ```

2. Download SQuAD Data.
   a. Download training data.
      ```
      > wget https://rajpurkar.github.io/SQuAD-explorer/dataset/train-v1.1.json
      ```
   b. Download development data.
      ```
      > wget https://rajpurkar.github.io/SQuAD-explorer/dataset/dev-v1.1.json
      ```
   c. Move the training and development files into a new folder.
      ```
      > mkdir squad
      ```
      ```
      > mv train-v1.1.json dev-v1.1.json squad
      ```

3. Download the BERT-base-cased pre-trained language model.
   a. The model can be downloaded by the following command. Note that the model requires approximately 400M disk space, so please be patient to wait until it finishes.
      ```
      > wget https://storage.googleapis.com/bert_models/2018_10_18/cased_L-12_H-768_A-12.zip
      ```
   b. After downloading the model, you should unzip the file by the following command.
      ```
      > unzip cased_L-12_H-768_A-12.zip
      ```
   c. The newly created folder contains the following files, where the files with the prefix "bert_model" are the model files, "vocab.txt" is the vocabulary, and "bert_config.json" is the configuration file.
      ```
      cased_L-12_H-768_A-12/
         |- bert_model.ckpt.meta
         |- bert_model.ckpt.index
      ```

```
|- bert_model.ckpt.data-00000-of-00001

|- vocab.txt

|- bert_config.json
```

    d. Transfer model files to a Google Cloud Storage bucket (path start with "gs://"). In this protocol, we use "gs://temp-bucket" for illustration. Note that using the Google Cloud Storage bucket is mandatory for TPU computing. If you are using CPU/GPU, this step can be omitted.

```
> gsutil -m cp cased_L-12_H-768_A-12/* gs://temp-bucket/bert
```

Note: Although we use an English dataset and pre-trained language model (PLM) to illustrate the protocol, it is also applicable to other datasets and PLMs, such as CMRC 2018 (Cui et al., 2019) with Chinese BERT (Devlin et al., 2019), as long as they share the same dataset and PLM structure.

## Training an English MRC system

Timing: 30 min

In this step, we will train a typical English machine reading comprehension system by using SQuAD dataset and pre-trained BERT model. After the training is complete, we will use the official evaluation script to obtain the system's performance on how well it solves the natural questions.

4. Check the training script and fill in with proper values.
    a. Open the training script.
```
> vim train_squad.sh
```
    b. The training script contains the following variables and arguments.
```
GS_BUCKET=gs://your-bucket

TPU_NAME=your-tpu-name

TPU_ZONE=your-tpu-zone

MODEL_OUTPUT_DIR=$GS_BUCKET/path-to-output-dir

python -u run_squad.py \
   --vocab_file=$GS_BUCKET/bert/cased_L-12_H-768_A-12/vocab.txt \
   --bert_config_file=$GS_BUCKET/bert/cased_L-12_H-768_A-12/bert_config.json \
   --init_checkpoint=$GS_BUCKET/bert/cased_L-12_H-768_A-12/bert_model.ckpt \
   --do_train=True \
```

```
                --train_file=./squad/train-v1.1.json \
                --do_predict=True \
                --predict_file=./squad/dev-v1.1.json \
                --train_batch_size=64 \
                --predict_batch_size=32 \
                --num_train_epochs=3.0 \
                --max_seq_length=512 \
                --doc_stride=128 \
                --learning_rate=3e-5 \
                --version_2_with_negative=False \
                --output_dir=$MODEL_OUTPUT_DIR \
                --do_lower_case=False \
                --use_tpu=True \
                --tpu_name=$TPU_NAME \
                --tpu_zone=$TPU_ZONE
```
   c. Fill the variables with proper values.
      i. $GS_BUCKET: This is the path for Google Cloud Storage. As we indicated in the previous section, we use "gs://temp-bucket" here.
      ii. $TPU_NAME: This is the name of TPU, which was created by using "ctpu" or "gcloud compute" commands. We use "t2-1" here.
      iii. $TPU_ZONE: This is the zone of TPU, which was created by using "ctpu" or "gcloud compute" commands. We use "us-central1-f" here.
      iv. $MODEL_OUTPUT_DIR: This is the location where we wish to save our model files. We use "squad-model" here.
      v. Other parameters are set with default values, and there is no need to change them at this time.
   d. Save the changes and exit by pressing ESC and typing the following.

```
> :wq
```

5. Train an English MRC system by using the preset training script.
   a. Type the following command to allow the model training to be executed in the background and save the log file into "train.log".
   
   ```
   > nohup nice bash run.squad.sh &> train.log &
   ```

b. The training script will automatically process the data, pre-train the model and perform task fine-tuning. The whole process takes approximately 30 min. The training may unusually fail due to the TPU issue. Please see troubleshooting 3 for further illustration.

6. Evaluate the system performance with the official script.
    a. Retrieve the prediction file of the development set from the storage bucket to the current folder.

    ```
    > gsutil cp gs://temp-bucket/squad-model/predictions.json.
    ```

    b. Type the following command to obtain the performance of the MRC system.

    ```
    > python eval_squad.py squad/dev-v1.1.json predictions.json
    ```

    c. After running the evaluation script, the performance results will be shown. We can see that the exact match (EM) score is 80.567, and the F1 score is 88.117.

```
{"exact": 80.56764427625355, "f1": 88.11721947565059, "total": 10570,
"HasAns_exact": 80.56764427625355, "HasAns_f1": 88.11721947565059,
"HasAns_total": 10570}
```

Note: As the main goal of our previous work (Cui et al., 2022) was to provide robust and comprehensive analyses of machine reading comprehension models, we carried out each experiment five times with different random seeds, and their average scores were used. However, to minimize the training time, we only train one model in this protocol, and it can be easily generalized to multiple runs as well by running steps 1–3 multiple times. Additionally, please note that TensorFlow with GPU or TPU suffers from the indeterministic issue, even with a fixed random seed. Please see troubleshooting 4 for further illustration.

## Collecting data points

Timing: 2 h

In this step, we will collect the data points for visualizations. We will mask each attention zone (4 zones and masking all, resulting in 5 in total) in each transformer layer (12 in total) and obtain their prediction files (5×12=60 in total).

7. Decode SQuAD development set multiple times to obtain the prediction files when disabling each attention zone in the different layers.
    a. Open the decoding script.

    ```
    > vim decode_squad.sh
    ```

    b. The decoding script is similar to the training script, with two additional arguments, "--mask_layer" and "--mask_zone" specified. "--mask_layer" indicates the layer to be masked, where the index ranges from 0 to 11 (12 layers in total). If "None" is set, it means that all layers will be masked. "--mask_zone" indicates the attention zone to be masked, where the valid

values are "q2", "q2p", "p2q", "p2", and "all". The final decoding script is as follows.

```
GS_BUCKET=gs://your-bucket

TPU_NAME=your-tpu-name

TPU_ZONE=your-tpu-zone

MODEL_OUTPUT_DIR=$GS_BUCKET/path-to-output-dir

for idx in {0..11};

do

python -u run_squad.py \
   --vocab_file=$GS_BUCKET/bert/cased_L-12_H-768_A-12/vocab.txt \
   --bert_config_file=$GS_BUCKET/bert/cased_L-12_H-768_A-12/bert_config.json \
   --init_checkpoint=$GS_BUCKET/bert/cased_L-12_H-768_A-12/bert_model.ckpt \
   --do_train=False \
   --do_predict=True \
   --predict_file=./squad/dev-v1.1.json \
   --predict_batch_size=32 \
   --max_seq_length=512 \
   --doc_stride=128 \
   --mask_layer=$idx \
   --mask_zone="q2" \
   --output_dir=$MODEL_OUTPUT_DIR \
   --do_lower_case=False \
   --use_tpu=True \
   --tpu_name=$TPU_NAME \
   --tpu_zone=$TPU_ZONE
done
```

      c.    By changing the values in "--mask_zone", we will obtain all prediction files (60 in total).

      d.    The prediction files will be saved in the "output_dir". The name will look like "predictions_layer0_q2.json", indicating its layer number and masked attention zone.

8. Collect the prediction files and evaluate their performances.

      a.    Copy the prediction files from the storage bucket to the local file system.

```
> mkdir prediction && cd prediction

> gsutil cp gs://temp-bucket/squad-
model/predictions_layers*.json .
```

      b.    We get the performances in each prediction file and write them into a result file (results.csv). Make sure the prediction files are in the "prediction" folder as created in the previous step. In this protocol, we use the exact match (EM) score as the source of data points, while we can also use the F1 score for visualization, which can be simply changed by passing an additional argument "--use-f1" to the "get_results.py" script.

```
> python get_results.py prediction results.csv
```

      c.    The following is a snippet of the result file (results.csv), where each row contains the result of each masked attention zone. For example, the third line represents the results in layer 2, yielding 80.028, 80.114, 80.36, 80.388, and 76.424 in all, $Q^2$, Q2P, P2Q, and $P^2$ zones, respectively.

```
layer,all,q2,q2p,p2q,p2
1,78.344,80.539,80.482,79.991,66.982
2,80.028,80.114,80.36,80.388,76.424
3,79.688,80.227,80.293,79.792,78.666
......
```

Note: The decoding time can be saved significantly by using multiple computing devices. For example, one can simultaneously decode each attention zone in step 4.

Note: The layer index starts from 0 in the pre-trained language model, indicating the first transformer layer. In this protocol, we directly use "layer 1" to indicate the first layer to avoid confusion.

### Visualizing effect of each attention zone

Timing: 5 min

In this step, we will use the result file to visualize the effect when a specific attention zone is disabled.

9. Decode SQuAD development set multiple times to get the prediction files when disabling each attention zone in the different layers.
    a. Run the visualization script by passing the result file as the first argument, the baseline performance as the second argument, and the name of the output figure as the third argument. In this example, the result file is "results.csv", the baseline performance (EM) is "80.567" (refer to step 6.c), and the output figure name is "squad.pdf".

    ```
    > python visualize_att_zone.py results.csv 80.567 squad.pdf
    ```

    b. The visualization figure will be saved in the "squad.pdf" file, which is shown as follows.

## Expected outcomes

After a series of preceding steps, we can get the visualization result of each attention zone in the different layers, which is depicted in Figure 1A. We also include the original figure presented in our previous study (Cui et al., 2022) for comparison.

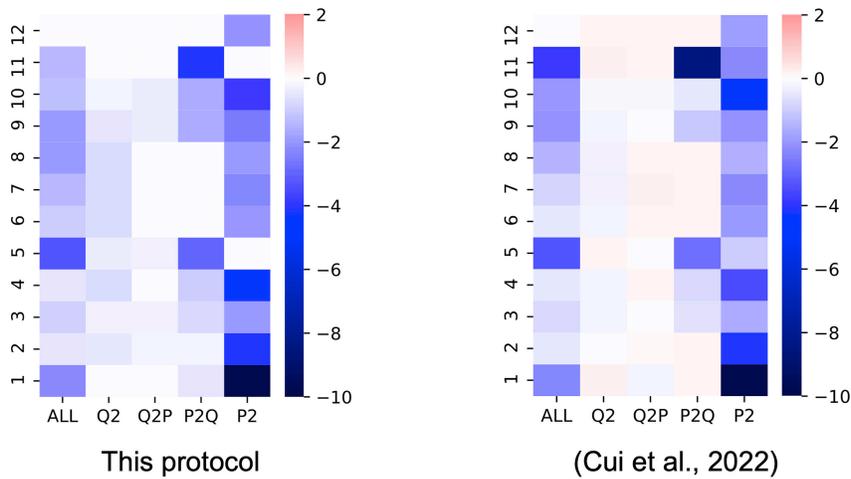

Figure 1. Visualization of attention zones in different layers using EM scores

(A) Visualization using the results in this protocol.

(B) Visualization presented in our previous study (Cui et al., 2022).

We can see that the two figures look similar to each other. The figure presented in this protocol (Figure 1A) is based on a single-run experiment, while our previous study used a five-run average as the source of visualization (Figure 1B). Overall, the P2Q and $P^2$ zones yield darker colors, indicating that removing the attention in these zones will result in worse system performance than $Q^2$ and Q2P zones.

A major difference between the two figures is that the color of Figure 1A is relatively lighter than that of Figure 1B. This will result in some analytical conclusions presented in our previous study not being drawn from Figure 1A. For example, in the 12th layer of Figure 1A, $Q^2$, Q2P, and P2Q show light red colors in Figure 1B, indicating that removing these attention zones will improve the system performance, which was verified by our previous work. However, we cannot observe this phenomenon in Figure 1A, yielding all white colors in these attention zones. This demonstrates that it is essential to carry out multiple experiments for reliable analytical studies. Further quantification and statistical analysis are shown in the next section.

One may wonder whether visualizing F1 scores will result in a completely different attention pattern. In this context, we also visualize the attention zones by using F1 scores instead of EM scores. This can be achieved by passing an additional argument "--use-f1" to the "get_results.py" script in step 8.b and changing the baseline score to "88.117" (refer to step 6.a) in step 9.a. The visualization is shown in Figure 2. We can see that Figure 2 has almost the same attention distribution as in Figure 1A, and thus, we only used EM scores for visualization for simplicity.

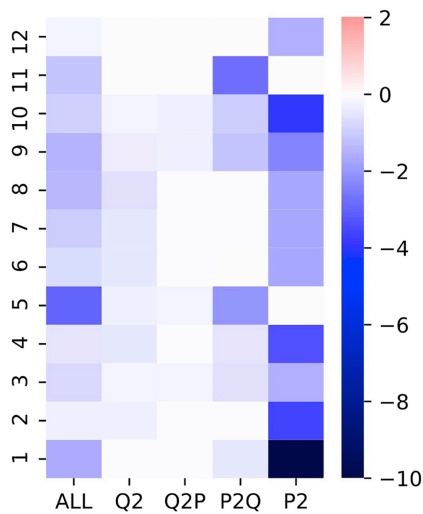

Figure 2. Visualization of attention zones in different layers using F1 scores

## Quantification and statistical analysis

In this protocol, we perform a single experiment to visualize the effect of each attention zone in different transformer layers. However, in our previous study (Cui et al., 2022), to ensure that the analyses were more robust and reliable, each experiment was performed five times, and their average scores were used for visualization and analyses. To explicitly understand the underlying reasons behind multiple runs, we show the original raw results as follows. This also helps in reproducing and comparing our results.

The single-run experimental result is shown in Table 1, which reflects the content of "results.csv" in step 8.c. These results were used for visualization, as shown in Figure 1A. The original results of Figure 1B are shown in Table 2, which is the exact figure that was shown in our previous study (Cui et al., 2022), where each experiment was performed five times and their average scores were reported. We also calculate the standard deviation between Table 1 and (Cui et al., 2022) for each attention zone in the last row of Table 2. The standard deviation for a specific attention zone is calculated by calculating the difference between Table 1 and (Cui et al., 2022) in a single layer, and then we calculate the standard deviation of these differences in all layers.

*Table 1. The results of masking each attention zone in different layers (single run)*

| Layer | All | $Q^2$ | Q2P | P2Q | $P^2$ |
|---|---|---|---|---|---|
| 1 | 78.344 | 80.539 | 80.482 | 79.991 | 66.982 |
| 2 | 80.028 | 80.114 | 80.360 | 80.388 | 76.424 |
| 3 | 79.688 | 80.227 | 80.293 | 79.792 | 78.666 |
| 4 | 80.028 | 79.896 | 80.577 | 79.527 | 76.291 |
| 5 | 77.237 | 80.180 | 80.227 | 77.512 | 80.568 |
| 6 | 79.536 | 79.886 | 80.539 | 80.568 | 78.553 |
| 7 | 79.149 | 79.858 | 80.539 | 80.492 | 78.231 |
| 8 | 78.600 | 79.858 | 80.464 | 80.539 | 78.647 |
| 9 | 78.666 | 79.991 | 80.218 | 78.978 | 78.061 |
| 10 | 79.347 | 80.407 | 80.180 | 78.997 | 76.670 |
| 11 | 79.158 | 80.577 | 80.549 | 76.405 | 80.568 |
| 12 | 80.539 | 80.568 | 80.568 | 80.501 | 78.505 |

*Table 2. The results for our previous study in comparison with Table 1*

| Layer | All | $Q^2$ | Q2P | P2Q | $P^2$ |
|---|---|---|---|---|---|
| 1 | 78.397 | 80.924 | 80.466 | 80.840 | 66.275 |
| 2 | 80.233 | 80.609 | 80.795 | 80.874 | 76.519 |
| 3 | 79.890 | 80.513 | 80.679 | 80.085 | 79.031 |
| 4 | 80.261 | 80.445 | 80.880 | 79.959 | 77.223 |
| 5 | 77.377 | 80.855 | 80.753 | 77.925 | 79.709 |
| 6 | 80.238 | 80.451 | 80.861 | 80.869 | 78.704 |
| 7 | 79.828 | 80.390 | 80.907 | 80.848 | 78.458 |
| 8 | 79.256 | 80.416 | 80.865 | 80.872 | 79.156 |
| 9 | 78.632 | 80.526 | 80.672 | 79.599 | 78.575 |
| 10 | 78.679 | 80.539 | 80.536 | 80.267 | 76.333 |
| 11 | 76.846 | 80.935 | 80.889 | 72.157 | 78.503 |
| 12 | 80.761 | 80.901 | 80.901 | 80.850 | 78.880 |
| Standard deviation of difference | 0.819 | 0.152 | 0.131 | 1.405 | 0.815 |

Through the comparisons between Tables 1 and 2, we can see that the standard deviation varies greatly among different attention zones. The less important attention zones $Q^2$ and Q2P yield minor variance, while the P2Q and $P^2$ zones yield larger variance, resulting in the standard deviations of 1.405 and 0.815, respectively. This is why multiple-run experiments are needed, especially for analytical studies, where the majority of the analyses stem from these results.

## Limitations

This protocol provides a step-by-step description of building an English machine reading comprehension system, collecting data points, and visualizing behaviors in different attention zones and layers. Our previous study also performed experiments on Chinese datasets and with different pre-trained language models, such as ELECTRA (Clark et al., 2020), etc. The protocol has good generalizability in these settings but may be slightly different in other settings, such as using the dataset in different datasets and pre-trained language models. However, such modifications are estimated to be minor, as the main technique presented is relatively independent of language and model type.

## Troubleshooting

### Problem 1

Using this protocol under a different operating system, such as macOS or Windows.

### Potential solution

Although our protocol is under the Linux system with the TPU computing device, it is also applicable to other operating systems (OS) with different compute devices (CPU or GPU). Here we provide several tips on migrating our protocol to these settings.

Some commands can be adapted to a different OS. For example, "wget" command in Linux can be changed with "curl" in macOS. For Windows users, we can also directly perform file management in a graphical interface without using command lines.

macOS is equipped with Python, and most commands in this protocol can be used. For Windows users, one can easily install Python and other dependencies via the online tutorial: https://packaging.python.org/en/latest/tutorials/installing-packages.

If you are using CPU or GPU for training models, please set "--use_tpu" argument as "False" in steps 4.b and 7.b and ignore the other arguments that are related to TPU.

### Problem 2

Using the "pip" command to install Python libraries is recommended in most cases, as it will also automatically install any related dependencies without seeking them individually. Sometimes, we may encounter a common error message as follows (step "package installation" in "before you begin" section).

```
ERROR: Could not install packages due to an EnvironmentError: [Errno 13]
Permission denied: '/some-path-to-library'
```

### Potential solution

This problem is a common issue when using the "pip" command, which indicates that we do not have enough permissions to install packages. To solve this problem, we can simply pass an additional argument "--user" to the "pip" command, such as follows (suppose we are installing "seaborn" package).

```
> pip install seaborn --user
```

If the problem persists, we can also use "sudo" command to allow us to access restricted files and operations. After executing the following command, we should also input the password to complete this command. Please make sure that the current user has "sudo" permissions. If you are not in "sudo" group, please contact your administrator to grant you "sudo" permissions.

```
> sudo pip install seaborn
```

### Problem 3

The training log shows an error message, and the training fails (step 5).

```
Resource exhausted: Attempting to reserve 14.10G at the bottom of memory.
That was not possible. There are 14.32G free, 0B reserved, and 14.09G
reservable.
```

### Potential solution

Although this error message may not occur with our training protocol and hyperparameter settings, it may rarely occur when the TPU server is extremely busy, and the TPU device may sometimes fail. If this happens, we recommend doing one of the following steps to resolve this issue.

Try to use a smaller batch size in step 4.b. Note that the batch size should be a multiple of 8.

If the error persists, we recommend deleting the TPU and adding a new TPU instead. This will resolve most of the issues related to unexpected TPU errors.

Further common issues related to the use of TPU can also be found at https://cloud.google.com/tpu/docs/troubleshooting/trouble-tf.

### Problem 4

When using a fixed random seed, the training results are not deterministic (note in step 6).

### Potential solution

It is a well-known issue that TensorFlow with GPU or TPU computing device suffers from the nondeterministic issue, even with a fixed random seed. This means that a fixed random seed will not lead to the same training result at the end. This is a complex issue that is related to many possible reasons, including different versions of libraries, different computing architectures, different parallelism strategies, etc. Thus, we recommend running each experiment multiple times and obtaining their average scores for visualization to ensure that the results are more robust.

It is worth noting that although some libraries, such as PyTorch (Paszke et al., 2019), do not seem to suffer from this issue, we also recommend running multiple experiments to achieve robust and stable results using different random seeds, as the weight initialization (which depends on the chosen random seed) is critical for training deep learning models.

# Article info

## Resource availability

### Lead contact

Further information and requests for resources and reagents should be directed to and will be fulfilled by the lead contact, Yiming Cui (ymcui@ir.hit.edu.cn).

### Materials availability

This study did not generate new unique reagents.

### Data and code availability

The datasets and code are available at GitHub: https://github.com/ymcui/mrc-model-analysis and archived at Zenodo with a https://doi.org/10.5281/zenodo.6522993.

## Acknowledgments

Yiming Cui would like to thank Google TPU Research Cloud (TRC) program for Cloud TPU access. This work is supported by Natural Science Foundation of Heilongjiang Province, China (Grant No. YQ2021F006).

## Author contributions

Conceptualization, Y.C.; methodology, Y.C.; formal analysis, Y.C. and W.Z.; writing – original draft, Y.C.; writing – review & editing, Y.C., W.Z., and T.L.; visualization, Y.C.; supervision, T.L.

## Declaration of interests

The authors declare no competing interests.